\documentclass{article}

\usepackage[final,nonatbib]{neurips_2023}

\usepackage[utf8]{inputenc} 
\usepackage[T1]{fontenc}    
\usepackage{hyperref}       
\usepackage{url}            
\usepackage{booktabs}       
\usepackage{amsfonts}       
\usepackage{nicefrac}       
\usepackage{microtype}      
\usepackage{xcolor}         

\usepackage{hyperref}
\usepackage{amsmath}    
\usepackage{multirow}
\usepackage{graphicx}
\usepackage{makecell}
\usepackage{pifont}

\usepackage{algorithm}
\usepackage[noend]{algorithmic}

\title{SAD: Segment Any RGBD}

\author{
  Jun Cen$^{1,3}$ \quad Yizheng Wu$^{2,3}$ \quad Kewei Wang$^{2,3}$ \quad Xingyi Li$^{2,3}$ \quad Jingkang Yang$^{3}$ \\ \textbf{Yixuan Pei}$^{4}$ \quad \textbf{Lingdong Kong}$^{5}$ \quad \textbf{Ziwei Liu}$^{3}$ \quad \textbf{Qifeng Chen}$^{1}$\\
  \\
The Hong Kong University of Science and Technology$^1$ \\
Huazhong University of Science and Technology$^2$ \\
Nanyang Technological University$^3$ \\
  Xi'an Jiaotong University$^4$ \\
National University Singapore$^5$
}

\begin{document}

\maketitle

\begin{figure}[h]
	\begin{center}
		\includegraphics[width=1.0\linewidth, height=0.5\linewidth]{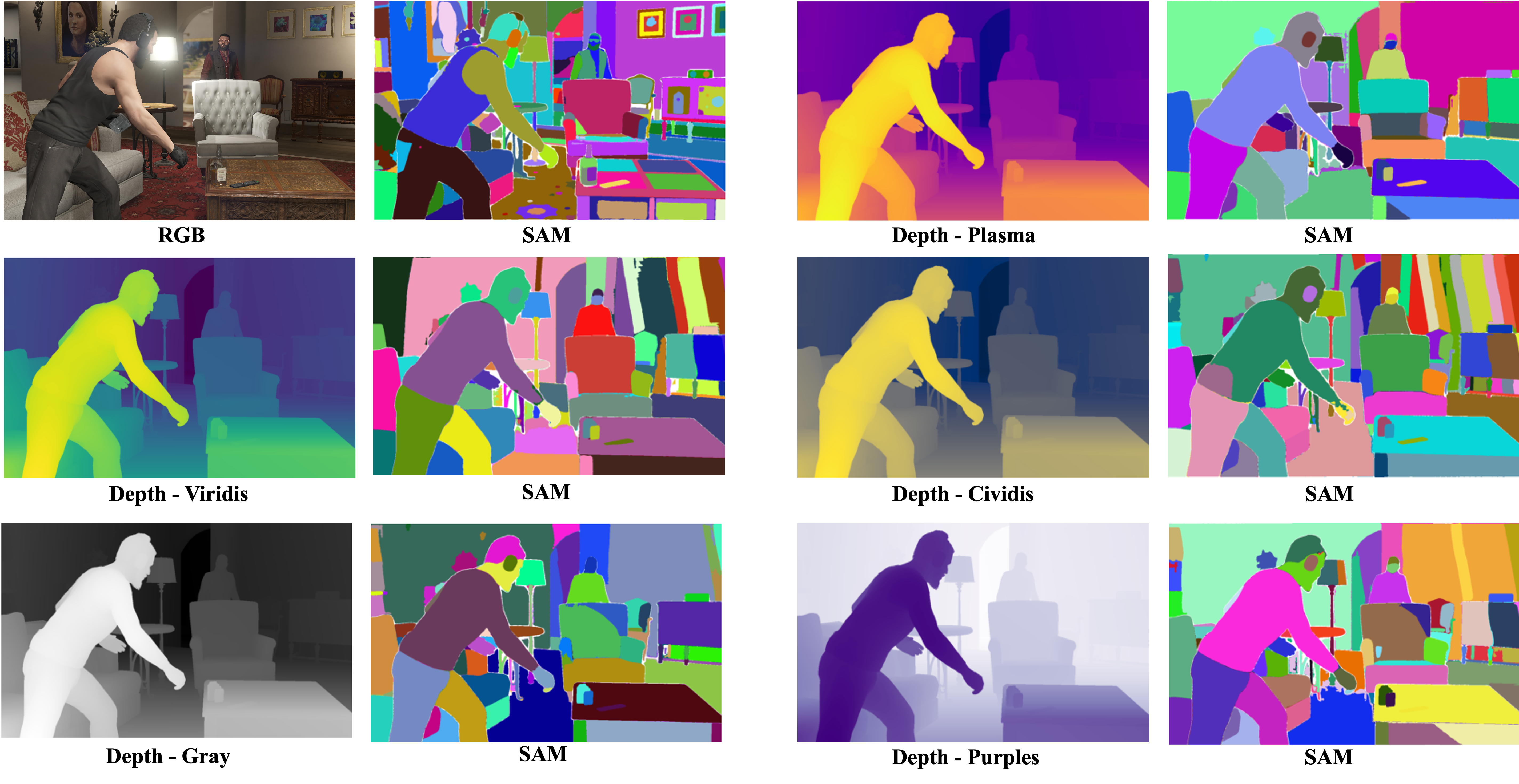}
         \vspace{-10pt}
		\caption{\textbf{Segmentation results of depth map by SAM.} In contrast to RGB images, segmentation results derived from depth maps inherently encompass a richer set of geometric information.}
        \vspace{-10pt}
		\label{fig:example}
	\end{center}
\end{figure}

\begin{abstract}

The Segment Anything Model (SAM) has demonstrated its effectiveness in segmenting any part of 2D RGB images. However, SAM exhibits a stronger emphasis on texture information while paying less attention to geometry information when segmenting RGB images. To address this limitation, we propose the Segment Any RGBD (SAD) model, which is specifically designed to extract geometry information directly from images. Inspired by the natural ability of humans to identify objects through the visualization of depth maps, SAD utilizes SAM to segment the rendered depth map, thus providing cues with enhanced geometry information and mitigating the issue of over-segmentation. We further include the open-vocabulary semantic segmentation in our framework, so that the 3D panoptic segmentation is fulfilled.
The project is available on \href{https://github.com/Jun-CEN/SegmentAnyRGBD}{https://github.com/Jun-CEN/SegmentAnyRGBD}.

\end{abstract}
\section{Introduction}
Recently, a prominent model for image segmentation called the Segment Anything model (SAM) has been introduced~\cite{kirillov2023segment}. SAM serves as a robust foundation model for effectively segmenting 2D images in various scenarios.
However, during the segmentation of 2D RGB images, it has been observed that SAM primarily relies on texture information, such as color, leading to over-segmentation results. Consequently, the challenge lies in finding a way to obtain segmentation results that incorporate more geometric information through the utilization of SAM.

To address this issue, we draw inspiration from the remarkable ability of humans to identify objects by visualizing depth maps.
We first map a depth map ($\mathbb{R}^{H\times W}$) to the RGB space ($\mathbb{R}^{H\times W\times 3}$) by a colormap function and then feed the rendered depth image into SAM. Compared to RGB images, the rendered depth image ignores the texture information and focuses on the geometry information, as shown in Fig.~\ref{fig:example}. Notably, it is worth mentioning that while previous SAM-based projects~\cite{zhang2023comprehensive} like SSA~\cite{SSA}, Anything-3D~\cite{shen2023anything3d}, and SAM 3D~\cite{pointcept2023} primarily employ RGB images as inputs, we are the first to utilize SAM for directly segmenting rendered depth images.
\section{Method}

\begin{figure}[h]
	\begin{center}
		\includegraphics[width=1.0\linewidth]{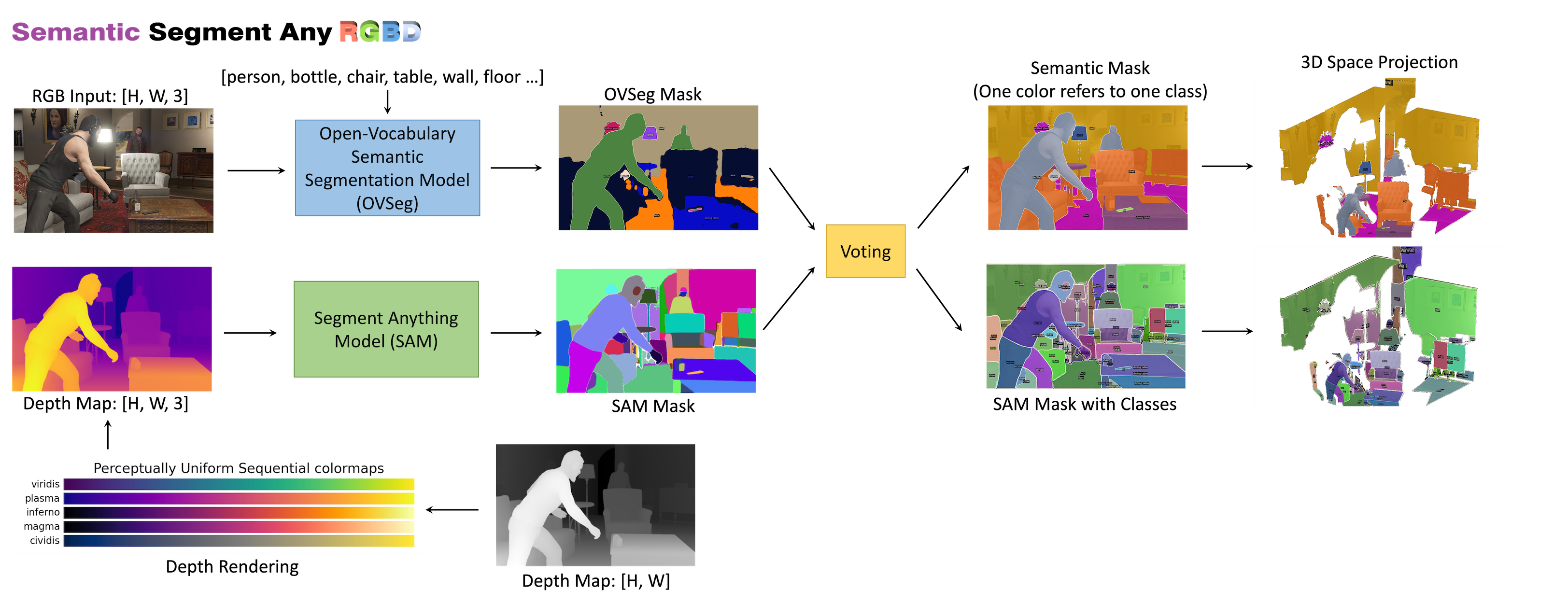}
         \vspace{-10pt}
		\caption{\textbf{The overview of the proposed SAD.} }
        \vspace{-10pt}
		\label{fig:flowchart}
	\end{center}
\end{figure}

\subsection{Preliminaries}
\textbf{Segment Anything Model (SAM).} The Segment Anything model (SAM)~\cite{kirillov2023segment} is a recently developed large Vision Transformer (ViT)-based model. SAM has been trained on an extensive visual corpus known as SA-1B. The training process on this large-scale dataset has endowed SAM with the remarkable ability to perform zero-shot segmentation on diverse styles of 2D images.

\textbf{Open-Vocabulary Semantic Segmentation (OVSeg).} Given the class candidates in the text format, Open-Vocabulary Semantic Segmentation (OVSeg)~\cite{liang2022open} can segment an image into semantic regions even if the categories are not seen during training.

\subsection{Segment Any RGBD}
In this paper, we propose Segment Any RGBD (SAD) that leverages both SAM and OVSeg to achieve semantic segmentation results utilizing the geometry information derived from depth maps.
The overview of SAD is shown in Fig.~\ref{fig:flowchart}. The process can be divided into the following parts:

\textbf{Rendering depth maps.} We notice that depth maps tend to emphasize geometry information over texture information when compared to RGB images, as visually depicted in Fig.~\ref{fig:example}. Capitalizing on this characteristic, our approach involves initially utilizing a colormap function~\cite{Hunter:2007} to render the depth maps to the RGB space. We try different colormaps such as Viridis, Gray, Plasma, Cividis, and Purples, as shown in Fig.~\ref{fig:example}. Consequently, the rendered depth maps are employed as inputs for SAM.

\textbf{Segmentation with SAM.} Following the rendering process, we apply SAM to the rendered depth images to generate initial SAM masks. It is worth noting that these initial SAM masks are class-agnostic and still over-segmented, as illustrated in the Fig.~\ref{fig:example}.

\textbf{Semantic segmentation with OVSeg.} By employing RGB images as input and leveraging text prompts, OVSeg exhibits the ability to generate coarse masks that encompass significant semantic information. These coarse semantic segmentation masks serve a dual role: firstly, they assist in guiding the clustering process of the over-segmented parts within the SAM masks, and secondly, they provide crucial category insights that contribute to refining the fine-grained SAM results.

\textbf{Semantic voting.} For each pixel in the SAM mask, we first find its corresponding predicted class from the OVSeg mask. Subsequently, we assign the class of each segment based on the majority class of pixels contained within it.
Following this, we can proceed to cluster adjacent segments that belong to the same class.

Finally, the semantic segmentation results can be projected to 3D-world based on the depth map for stereoscopic visualizations. This projection enables a comprehensive understanding and visual representation of the segmented results in their spatial context.

\section{Comparison with RGB Image Input}
\begin{figure}[t]
	\begin{center}
		\includegraphics[width=1.0\linewidth]{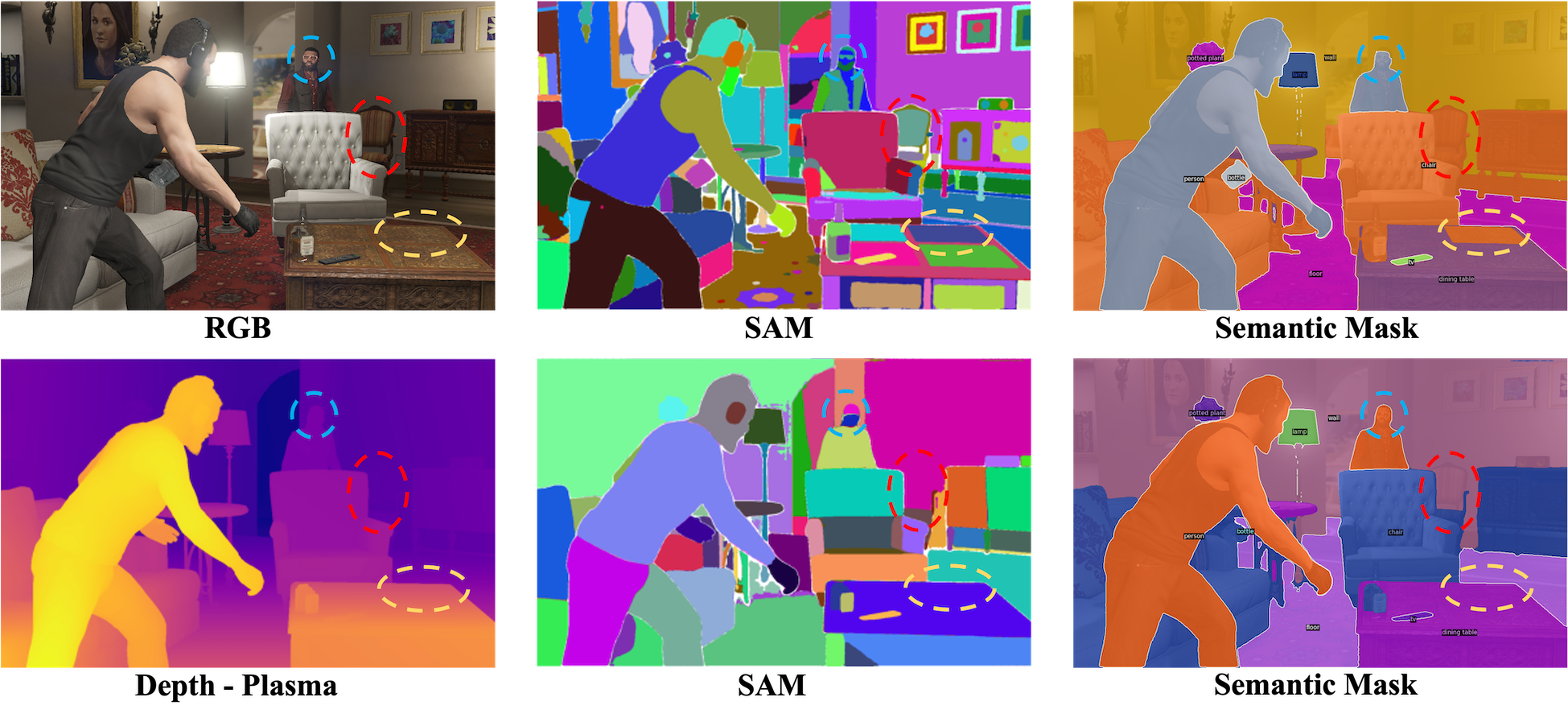}
         \vspace{-10pt}
		\caption{\textbf{Segmentation results with the RGB image input and the rendered depth image inputs.} }
        \vspace{-10pt}
		\label{fig:comparison}
	\end{center}
\end{figure}

We compare the proposed rendered depth image input with the RGB image input.
RGB images predominantly capture texture information, while depth images primarily contain geometry information. As a result, RGB images tend to be more vibrant and colorful compared to the rendered depth images. Consequently, SAM produces a larger number of masks for RGB inputs compared to depth inputs, as illustrated in Fig.~\ref{fig:comparison}.
The utilization of rendered depth images mitigates the issue of over-segmentation in SAM. For instance, when examining the table, the RGB images segment it into four distinct parts, with one part being misclassified as a chair in the semantic results (indicated by yellow circles in Fig.\ref{fig:comparison}), while it is accurately classified in the depth image.
It is also important to note that when two objects are in close proximity, they may be segmented as a single object in the depth image, as depicted by the chair in the red circle of Fig.~\ref{fig:comparison}. In such cases, the texture information present in RGB images becomes crucial for accurately identifying and distinguishing the objects.

\section{Qualitative Results}
\begin{figure}[h!]
	\begin{center}
		\includegraphics[width=1.0\linewidth]{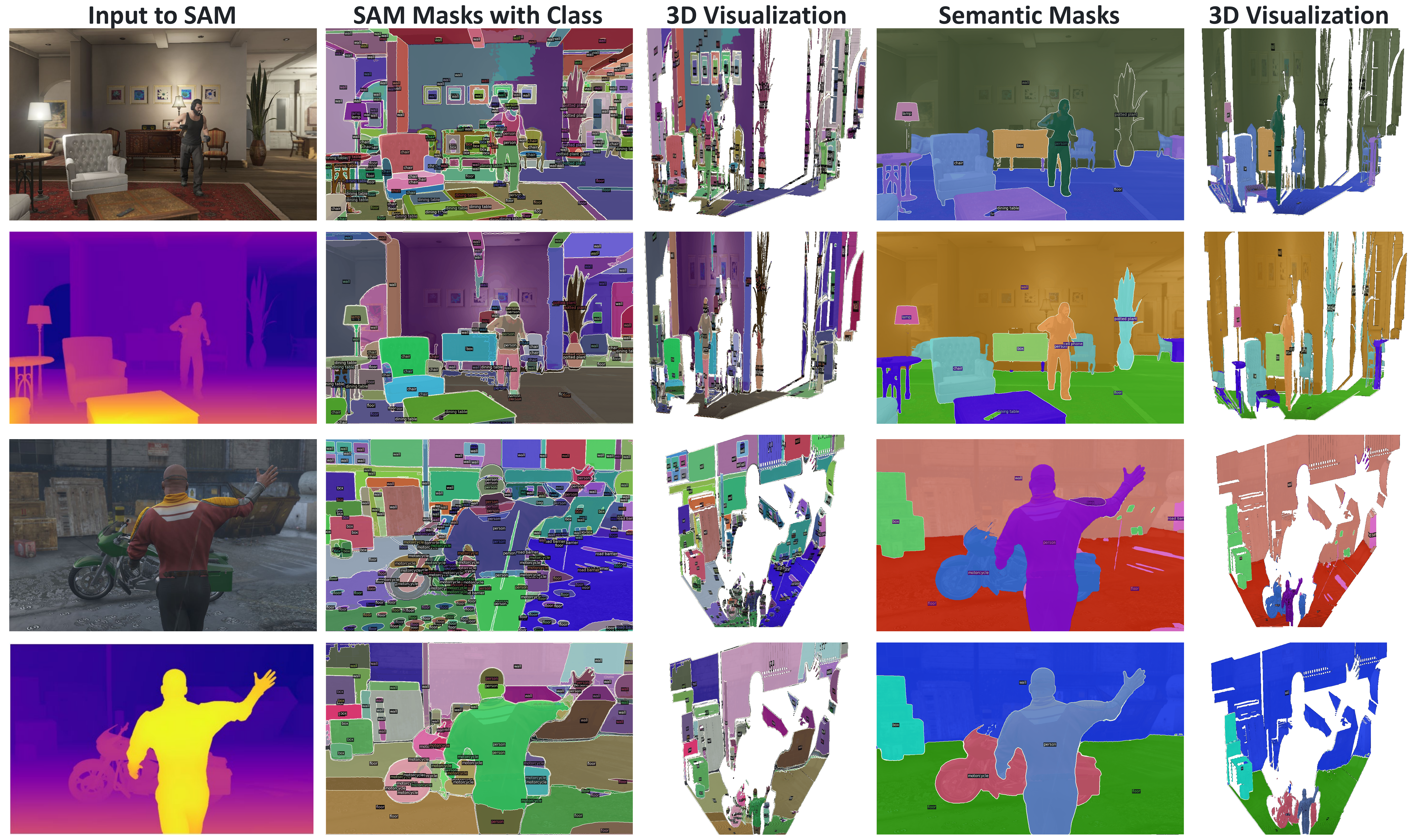}
         \vspace{-10pt}
		\caption{\textbf{Qualitative results on the Sailvos3D dataset.}}
        \vspace{-10pt}
		\label{fig:sailvos3D}
	\end{center}
\end{figure}

\begin{figure}[h!]
	\begin{center}
		\includegraphics[width=1.0\linewidth]{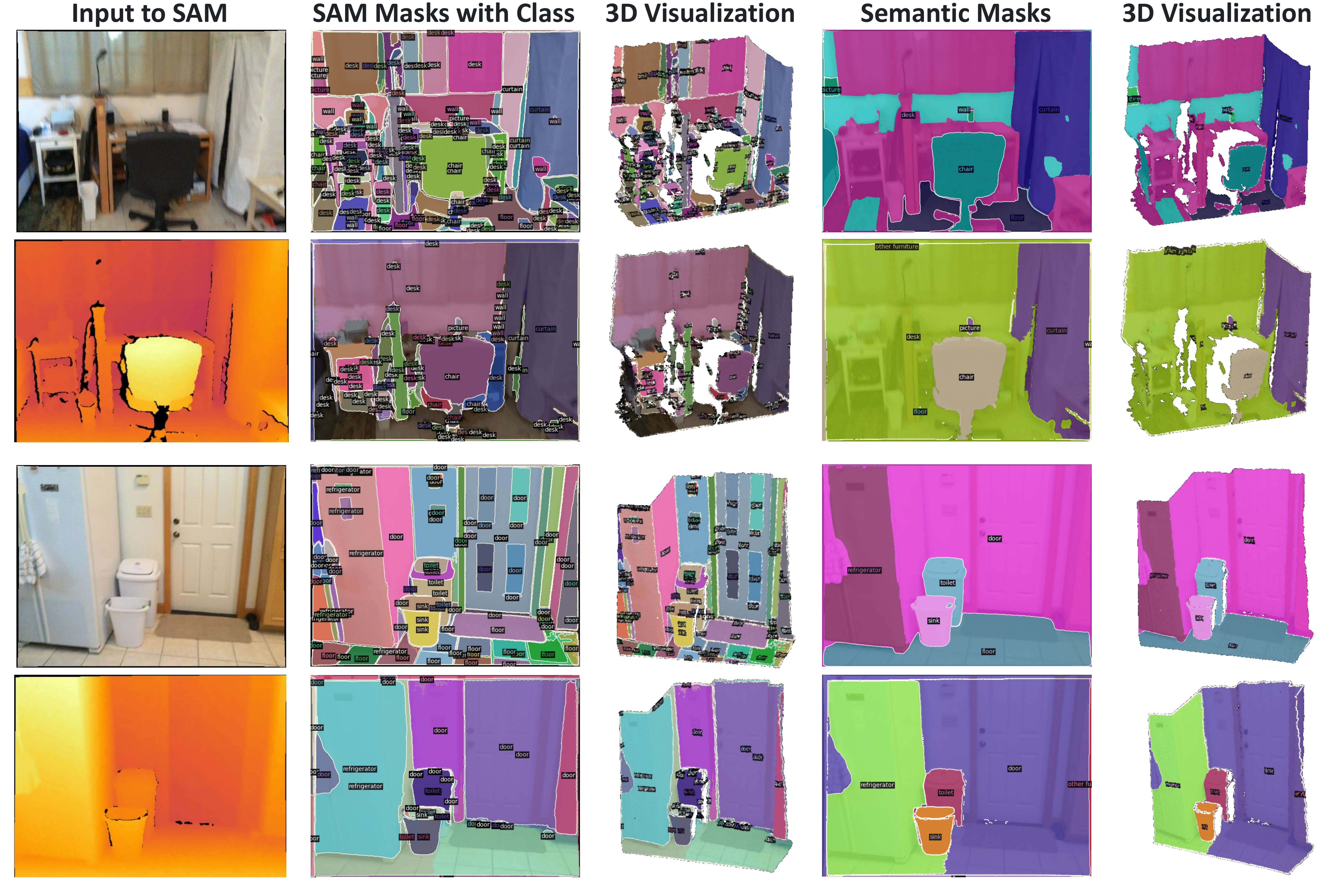}
         \vspace{-10pt}
		\caption{\textbf{Qualitative results on the ScannetV2 dataset.}}
        \vspace{-10pt}
		\label{fig:scannet}
	\end{center}
\end{figure}

We present qualitative results on Sailvos3D~\cite{HuCVPR2021} and ScanNet~\cite{dai2017scannet}, which are depicted in Fig.~\ref{fig:sailvos3D} and Fig.~\ref{fig:scannet}. The figures clearly demonstrate the enhanced performance of our method in generating geometric semantic segmentation results when utilizing depth map inputs.

\section{Conclusion}
In summary, we introduce the Segment Any RGBD (SAD) model, which combines SAM and OVSeg for semantic segmentation using depth maps. SAD leverages the geometry information of depth maps by rendering them with RGB information and feeding them into SAM. Initial SAM masks are generated, which are refined using OVSeg's coarse semantic segmentation masks.
The clustering process is then applied to group adjacent segments of the same class, improving the segmentation results' coherence. Finally, the semantic segmentation results are projected onto the 3D world based on the depth map, enabling comprehensive stereoscopic visualization.
Overall, SAD enhances semantic segmentation by incorporating depth maps and leveraging both SAM and OVSeg, resulting in more accurate and context-aware segmentation results. This work opens up new possibilities for advancing semantic segmentation tasks and provides valuable insights into real-world applications.

\bibliographystyle{IEEEbib}
\bibliography{refs}

\begin{thebibliography}{1}\itemsep=-1pt

\bibitem{SSA}
Jiaqi Chen, Zeyu Yang, and Li Zhang.
\newblock Semantic segment anything.
\newblock \url{https://github.com/fudan-zvg/Semantic-Segment-Anything}, 2023.

\bibitem{pointcept2023}
Pointcept Contributors.
\newblock Pointcept: A codebase for point cloud perception research.
\newblock \url{https://github.com/Pointcept/Pointcept}, 2023.

\bibitem{dai2017scannet}
Angela Dai, Angel~X. Chang, Manolis Savva, Maciej Halber, Thomas Funkhouser,
  and Matthias Nie{\ss}ner.
\newblock Scannet: Richly-annotated 3d reconstructions of indoor scenes.
\newblock In {\em CVPR}, 2017.

\bibitem{HuCVPR2021}
Y.-T. Hu, J. Wang, R.~A. Yeh, and A.~G. Schwing.
\newblock {SAIL-VOS 3D: A Synthetic Dataset and Baselines for Object Detection
  and 3D Mesh Reconstruction from Video Data}.
\newblock In {\em CVPR}, 2021.

\bibitem{Hunter:2007}
J.~D. Hunter.
\newblock Matplotlib: A 2d graphics environment.
\newblock {\em Computing in Science \& Engineering}, 9(3):90--95, 2007.

\bibitem{kirillov2023segment}
Alexander Kirillov, Eric Mintun, Nikhila Ravi, Hanzi Mao, Chloe Rolland, Laura
  Gustafson, Tete Xiao, Spencer Whitehead, Alexander~C Berg, Wan-Yen Lo, et~al.
\newblock Segment anything.
\newblock {\em arXiv preprint arXiv:2304.02643}, 2023.

\bibitem{liang2022open}
Feng Liang, Bichen Wu, Xiaoliang Dai, Kunpeng Li, Yinan Zhao, Hang Zhang,
  Peizhao Zhang, Peter Vajda, and Diana Marculescu.
\newblock Open-vocabulary semantic segmentation with mask-adapted clip.
\newblock In {\em CVPR}, 2023.

\bibitem{shen2023anything3d}
Qiuhong Shen, Xingyi Yang, and Xinchao Wang.
\newblock Anything-3d: Towards single-view anything reconstruction in the wild.
\newblock {\em arXiv preprint arXiv:2304.10261}, 2023.

\bibitem{zhang2023comprehensive}
Chunhui Zhang, Li Liu, Yawen Cui, Guanjie Huang, Weilin Lin, Yiqian Yang, and
  Yuehong Hu.
\newblock A comprehensive survey on segment anything model for vision and
  beyond.
\newblock {\em arXiv preprint arXiv:2305.08196}, 2023.

\end{thebibliography}


\end{document}